\documentclass{article}

\PassOptionsToPackage{numbers, compress}{natbib}
\usepackage[preprint]{neurips_2023}

\usepackage{url}            % simple URL typesetting
\usepackage{booktabs}       % professional-quality tables
\usepackage{amsfonts}       % blackboard math symbols
\usepackage{nicefrac}       % compact symbols for 1/2, etc.
\usepackage{microtype}      % microtypography
\usepackage{xcolor}         % colors

\definecolor{cvprblue}{rgb}{0.21,0.49,0.74}

\usepackage{xspace}
\usepackage{multirow}

\usepackage{times}
\usepackage{amsmath}
\usepackage{amssymb}
\usepackage{bbding}
\usepackage{enumitem}
\usepackage{tablefootnote}
\usepackage{makecell}
\usepackage{tcolorbox}
\usepackage{subcaption}

\usepackage{hyperref}
\hypersetup{breaklinks=true,colorlinks,citecolor=cvprblue}

\usepackage{amsmath}
\usepackage{amsfonts}
\usepackage{amssymb}
\usepackage{multirow}
\usepackage{mathtools} 
\usepackage{verbatim}
\usepackage{anyfontsize}
\usepackage{microtype}

\definecolor{mygreen}{HTML}{3cb44b}
\definecolor{skyblue}{HTML}{beffff}
\definecolor{lightgreen}{HTML}{90ee90}

\usepackage{enumitem}
\setitemize{itemsep=10pt,topsep=0pt,parsep=0pt,partopsep=0pt}

\newcommand{\RN}[1]{%
	\textup{\lowercase\expandafter{\it \romannumeral#1}}%
}

\newcommand{\beq}{\vspace{0mm}\begin{equation}}
\newcommand{\eeq}{\vspace{0mm}\end{equation}}
\newcommand{\beqs}{\vspace{0mm}\begin{eqnarray}}
\newcommand{\eeqs}{\vspace{0mm}\end{eqnarray}}
\newcommand{\barr}{\begin{array}}
\newcommand{\earr}{\end{array}}

\usepackage{color, colortbl}
\definecolor{Gray}{gray}{0.93}

\usepackage{lipsum}

\newcommand\blfootnote[1]{%
  \begingroup
  \renewcommand\thefootnote{}\footnote{#1}%
  \addtocounter{footnote}{-1}%
  \endgroup
}

\usepackage{xcolor,amsmath}
\usepackage[linesnumbered,ruled,vlined]{algorithm2e}
\DontPrintSemicolon

\usepackage{color, colortbl}

\definecolor{emerald}{rgb}{0.31, 0.78, 0.37}
\definecolor{coralred}{rgb}{1.0, 0.25, 0.25}

\newcommand{\MyColorBox}[2][red]%
{%
    \settowidth{\Width}{#2}%
    \colorbox{#1}%
    {%      
        \raisebox{-\DepthReference}%
        {%
                \parbox[b][\HeightReference+\DepthReference][c]{\Width}{\centering#2}%
        }%
    }%
}

\definecolor{codegray}{gray}{0.9}

\usepackage{xcolor,amsmath}
\usepackage[linesnumbered,ruled,vlined]{algorithm2e}
\DontPrintSemicolon

\renewcommand{\KwSty}[1]{\textnormal{\textcolor{blue!90!black}{\ttfamily\bfseries #1}}\unskip}

\renewcommand{\ArgSty}[1]{\textnormal{\ttfamily #1}\unskip}
\SetKwComment{Comment}{\color{green!50!black}\# }{}

\newcommand{\var}{\texttt}
\newcommand{\FuncCall}[2]{\texttt{\bfseries #1(#2)}}
\SetKwProg{Function}{def}{:}{}

\SetKwProg{For}{for}{:}{}
\SetKwProg{If}{if}{:}{}

\newcommand{\fakeparagraph}[1]{\vspace{3mm}\noindent\textbf{#1}}

\newcommand{\system}{Ours\xspace}
\newcommand{\dataname}{\textsc{LVIS-Instruct4V}\xspace}

\newcommand*{\eg}{\emph{e.g.}\@\xspace}
\newcommand*{\etc}{\emph{etc}\@\xspace}

\newcommand*{\vs}{\emph{vs.}\@\xspace}

\definecolor{bluegray}{rgb}{0.4, 0.6, 0.8}
\definecolor{atomictangerine}{rgb}{1.0, 0.6, 0.4}
\definecolor{bananayellow}{rgb}{1.0, 0.88, 0.21}
\definecolor{amethyst}{rgb}{0.6, 0.4, 0.8}
\definecolor{newblue}{rgb}{0.45, 0.56, 0.80}
\definecolor{aigold}{RGB}{244,210, 1} 
\definecolor{aired}{RGB}{255,180,181}
\definecolor{defaultcolor}{gray}{0.9}

\newlength\savewidth

\title{To See is to Believe: Prompting GPT-4V for Better Visual Instruction Tuning}

\author{Junke Wang$^{1*}$ \ \ 
Lingchen Meng$^{1*}$ \ \ 
Zejia Weng$^{1}$ \ \  
Bo He$^{2}$ \ \ 
Zuxuan Wu$^{1\dagger}$ \ \  
Yu-Gang Jiang$^{1}$ \vspace{0.1in}\\
{$^1$Fudan University} \ \ {$^2$ University of Maryland} 
}

\begin{document}
\maketitle
\begin{abstract}
Existing visual instruction tuning methods typically prompt large language models with textual descriptions to generate instruction-following data. Despite the promising performance achieved, these descriptions are derived from image annotations, which are oftentimes coarse-grained. Furthermore, the instructions might even contradict the visual content without observing the entire visual context. To address this challenge, we introduce a fine-grained visual instruction dataset, \dataname, which contains $220K$ visually aligned and context-aware instructions produced by prompting the powerful GPT-4V with images from LVIS. 
Through experimental validation and case studies, we demonstrate that high-quality visual instructional data could improve the performance of LLaVA-1.5, a state-of-the-art large multimodal model, across a wide spectrum of benchmarks by clear margins. Notably, by simply replacing the LLaVA-Instruct with our \dataname, we achieve better results than LLaVA on most challenging LMM benchmarks, \eg, LLaVA$^w$ (76.7 \vs 70.7) and MM-Vet (40.2 \vs 35.4). We release our data and model at \url{https://github.com/X2FD/LVIS-INSTRUCT4V}.
\blfootnote{$^{*}$ Equal contributions.}
\blfootnote{$^{\dagger}$Corresponding author.}
\end{abstract}

\section{Introduction}
\label{sec:intro}
Recent years have witnessed the booming development of Large Language Models (LLM)~\cite{brown2020language,zhang2022opt,touvron2023llama,touvron2023llama2}, which represent a remarkable advancement in AI by offering unprecedented language understanding and reasoning capabilities. The great potential of LLMs motivates researchers in the computer vision community to build large multimodal models (LMM) on top of LLMs such that visual inputs can be effectively analyzed.

\begin{figure}[t]
  \centering
  \includegraphics[width=0.9\linewidth]{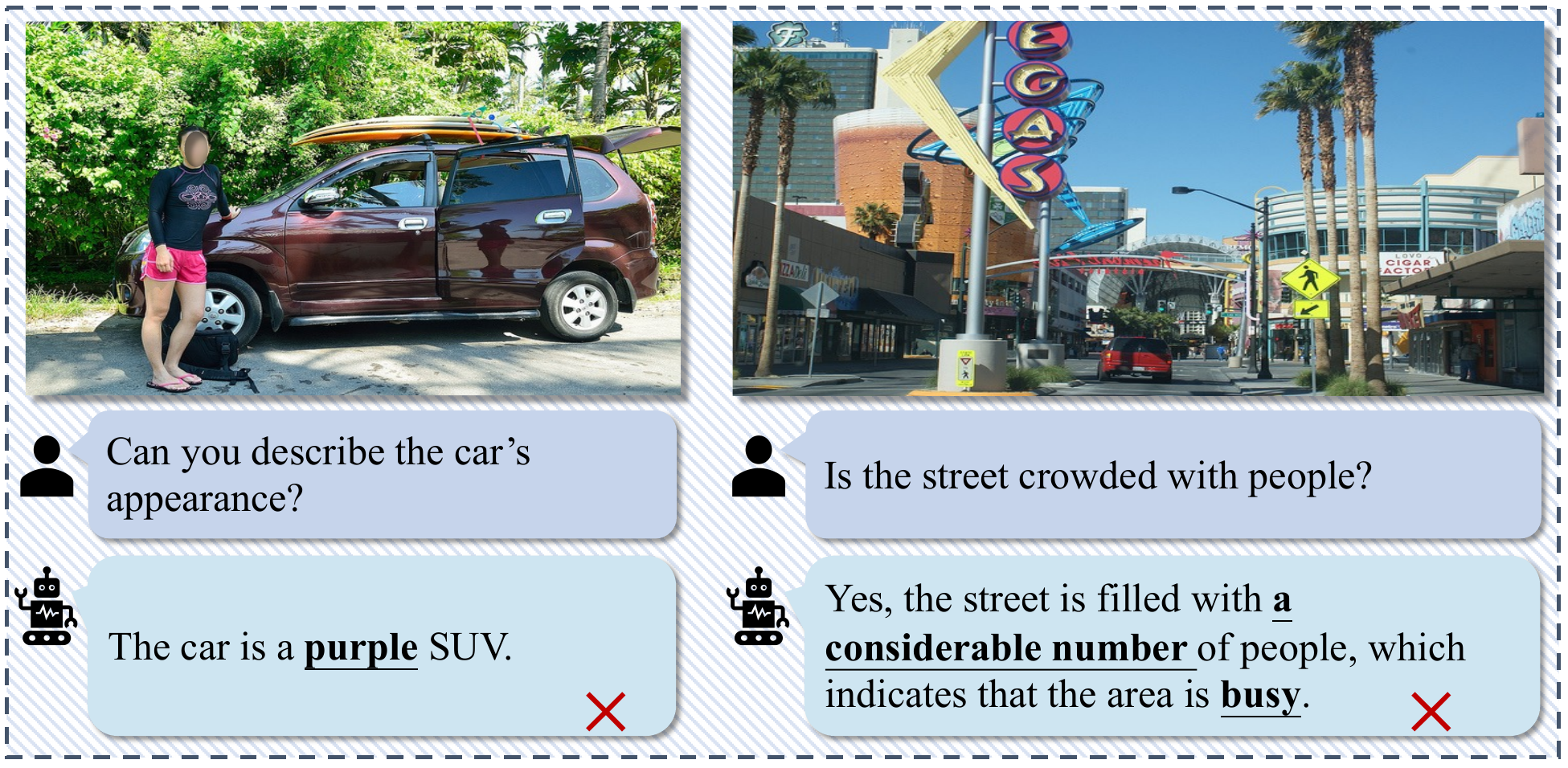}
  %\vspace{0.2in}
  \caption{Examples of incorrect instruction-answer pairs in LLaVA~\cite{li2023llava}, which are generated by prompting GPT-4. Without direct access to the images, GPT-4 may produce incorrect responses when asked fine-grained questions about color, location, \etc. }
  \label{fig:llava_wrong}
\end{figure}

Visual instruction tuning~\cite{liu2023visual} offers a promising way to bridge visual and textual information. Exemplary approaches, such as LLaVA~\cite{li2023llava,liu2023improved} and MiniGPT-4~\cite{zhu2023minigpt}, have demonstrated exceptional proficiency in natural instruction-following and visual reasoning. Typically, these methods leverage large language models~\cite{chatgpt,gpt4} to produce a large amount of instruction-answer pairs based on the textual description of the images, \eg, captions or bounding boxes with labels. In the absence of direct access to visual signals, crucial details are sometimes missing, and in some cases, the instructions are even incorrect, as depicted in Figure~\ref{fig:llava_wrong}.

To address this issue, we propose to harness GPT-4V, a powerful large multimodal model, to curate more accurate, context-aware, and fine-grained instruction-following data with a carefully designed pipeline. In particular, we feed GPT-4V with images from LVIS~\cite{gupta2019lvis}, an object detection dataset characterized by a comprehensive taxonomy and intricate annotations, along with their corresponding box annotations, and prompt the model to generate two types of instruction-answer pairs: contextually-aware conversations related to the visual contents and high-quality image descriptions. Finally, we have $220K$ detailed instructions, forming a large-scale instruction dataset, \dataname. 

\begin{figure}[t]
\centering
\includegraphics[width=0.65\linewidth]{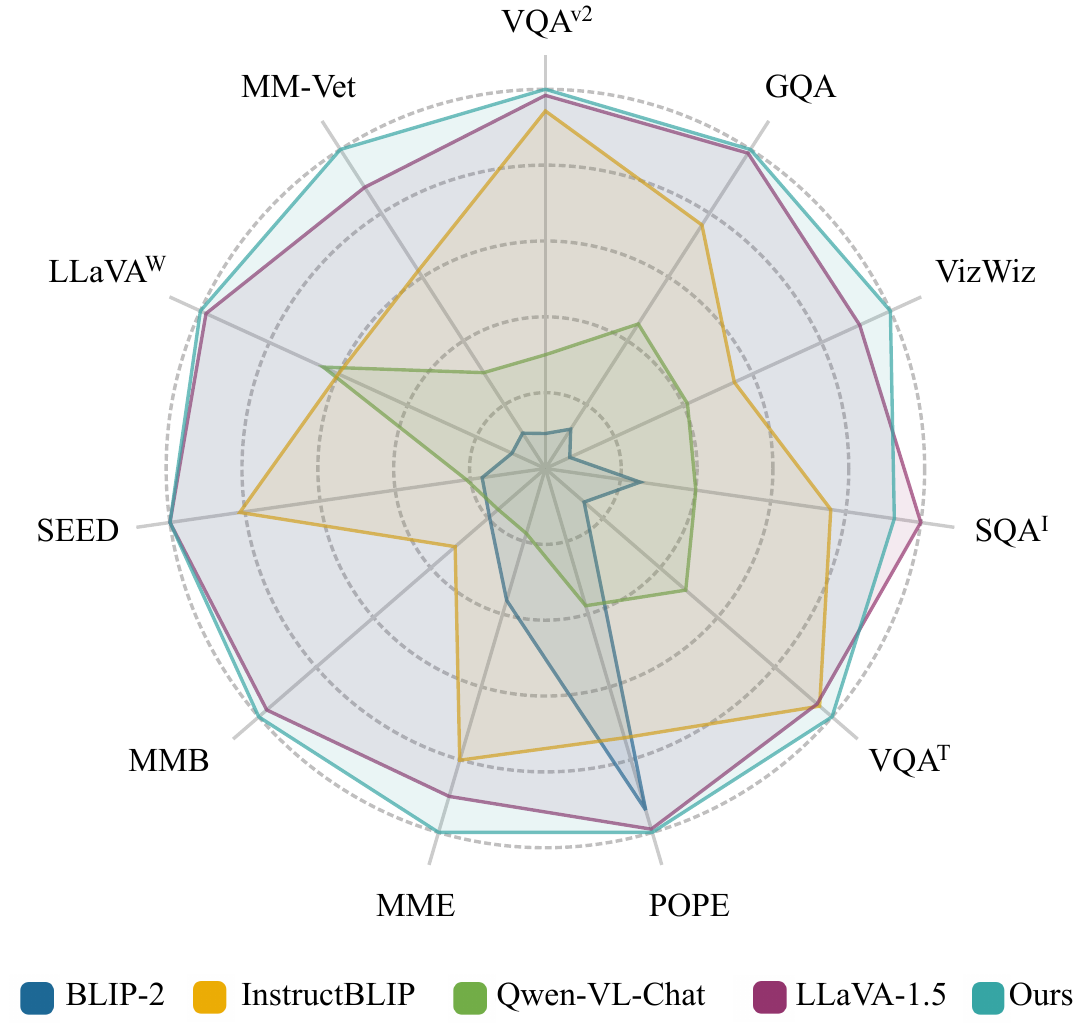}
\vspace{0.1in}
\caption{Comparisons of with other SOTA methods on various benchmarks. Please refer to Table~\ref{tab:benchmark} for detailed numbers.}
\label{fig:radar}
\end{figure}

We adopt the same model architecture with LLaVA-1.5~\cite{liu2023improved}, and replace their LLaVA-Instruct with our \dataname for visual instruction tuning. The experimental results demonstrate that significant improvements could be observed in most traditional QA benchmarks~\cite{goyal2017vqav2,gurari2018vizwiz} and more challenging LMM benchmarks, \eg, LLaVA$^w$ (76.7 \vs 70.7) and MM-Vet (40.2 \vs 35.4). In addition, the overall results could be further improved if we mix our data with LLaVA-Instruct (see Figure~\ref{fig:radar} for a better comparison with other state-of-the-art methods).

\section{Related Work}
\label{sec:related}
\subsection{Large Multimodal Models}
Large Multimodal Models (LMMs)~\cite{alayrac2022flamingo,chowdhery2022palm,li2023multimodal,li2023blip,driess2023palm}, also known as Multimodal Large Language Models (MLLMs), have attracted emerging attention due to their profound impact for a wide range of applications like image captioning~\cite{you2016image,yao2017boosting} and visual question-answering~\cite{antol2015vqa,shih2016look}. These methods are typically built upon Large Language Models (LLMs) to extend their powerful comprehension and reasoning capability to the multimodal domain. Frozen~\cite{tsimpoukelli2021multimodal} and Flamingo~\cite{alayrac2022flamingo} represent the pioneering attempts of LMMs, which are trained on interleaved visual and textual data for in-context few-shot learning. However, they directly feed visual tokens into language models without an intermediary projector, which presents a remarkable challenge for cross-modal alignment. To mitigate this issue, BLIP-2~\cite{li2023blip} proposes to connect the image encoder to a pretrained LLM through a lightweight Q-former. With this, two-stage pretraining is conducted to enforce the Q-Former to learn the textual grounded and LLM-interpretable visual representations, respectively.

\subsection{Visual Instruction Tuning} 
Instruction tuning~\cite{ouyang2022training,wang2022self,wang2022benchmarking} is first explored in the field of NLP, which finetunes or adaptes language models~\cite{brown2020language,raffel2020exploring,zhang2022opt,touvron2023llama} to follow specific textual instructions. Despite the prominent success in the NLP domain, the inherent text-only nature poses limitations on its ability to effectively comprehend and interact with visual content. To address this challenge, LLaVA~\cite{li2023llava} introduces \textit{visual instruction tuning} for the first time, enabling language models to follow visual instructions and engage with multimodal information. After that, a series of approaches have been developed to improve the visual instruction tuning with advanced architectures~\cite{idefics,dai2023instructblip,bai2023qwen} or more versatile functionalities~\cite{peng2023kosmos,zhu2023minigpt,lai2023lisa}. Nevertheless, the instruction data employed by these methods are generated either by manually reformating the existing annotated datasets like VQA~\cite{goyal2017vqav2}, or prompting GPT-4~\cite{gpt4} to derive from the textual description of images, which undoubtedly limits the fine-grained visual understanding of the results models. In this paper, we make use of the strong multimodal understanding capability of GPT-4V to collect a novel visual instruction dataset, \dataname. Benefiting from the direct access to images during the data generation process,  the instruction-answer pairs in \dataname are not only more visually aligned but also encompass rich visual details.

\section{LVIS-INSTRUCT4V}
\label{sec:method}

\fakeparagraph{Data collection.} The impressive success of Large Language Models (LLMs)~\cite{li2023blip,alayrac2022flamingo} has highlighted the potential of data scaling in training high-capacity models. However, the majority of existing multimodal datasets have predominantly been tailored for specific tasks, such as image-level understanding~\cite{cc,laion} or region-level referring comprehension~\cite{ref_coco,mdetr} While these datasets serve their respective purposes, they often lack the depth and complexity of data necessary to address more intricate challenges rooted in multi-turn interactions.

In contrast to prior task-specific designs, the acquisition of instruction-following data requires multi-turn interactions. Notably, LLaVA~\cite{liu2023visual} uses ChatGPT/GPT-4 to autonomously generate instruction-following data by leveraging location and semantics from detection data. However, the instructions are generated by a language-only approach using available annotations. This limitation can lead to the omission of crucial details or introduce inaccuracies when visual inputs are absent. 

\begin{figure}[t]
  \centering
  \includegraphics[width=\linewidth]{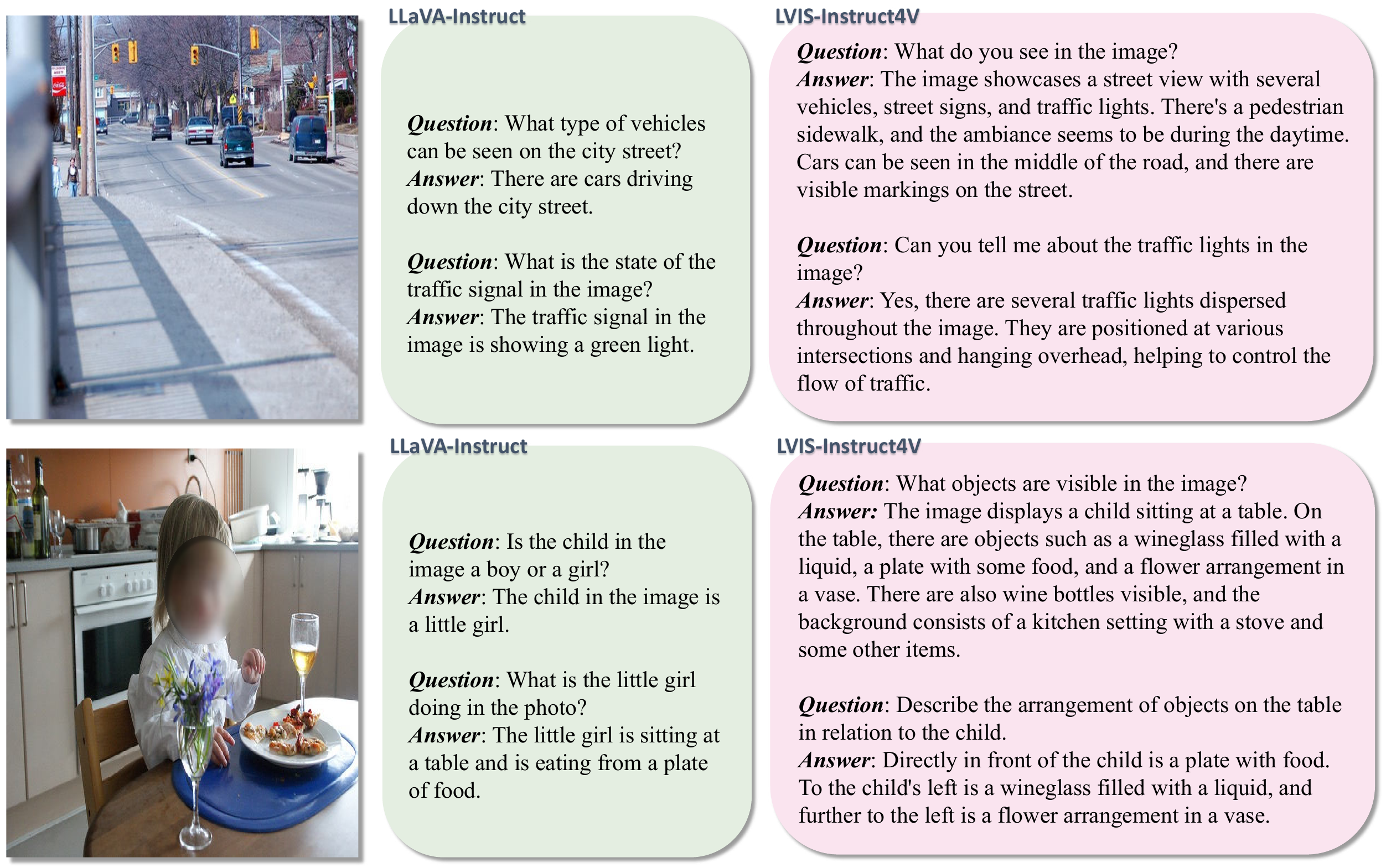}
  \vspace{-0.2in}
  \caption{Examples of the instruction-answer pairs from LLaVA and our instruction dataset.}
  \label{fig:com_vis}
  \vspace{-0.25in}
\end{figure}

In this paper, we aim to curate a fine-grained instruction-following dataset by explicitly conditioning the generation of instruction data on both language and visual inputs. This ensures that instructions are carefully crafted based on visual context, taking into account the categories and attributes of objects, as well as their interrelationships. To achieve this goal, we leverage image data from LVIS~\cite{gupta2019lvis}, as well as their fine-grained annotations of an extensive taxonomy~\footnote{Images in LVIS~\cite{gupta2019lvis} are all sourced from COCO, but the annotations are different.}, to prompt GPT-4V to undertake two key tasks: (1) generate conversational question-answer lists through self-reasoning and (2) produce high-quality image descriptions guided by precise bounding box information. For a comprehensive overview of the detailed prompts used in data generation, please refer to the Appendix~\ref{appendix:prompt}. To the end, we use $110K$ images from LVIS and generate $220K$ high-quality visual instructions, which consist of 110K conversational data and 110K descriptional data. Figure~\ref{fig:com_vis} gives an example of generated instructions and comparisons with those from LLaVA. We can observe that instructions based on visual and language inputs are more detailed.

\begin{figure}[!h]
  \centering
  \includegraphics[width=0.75\linewidth]{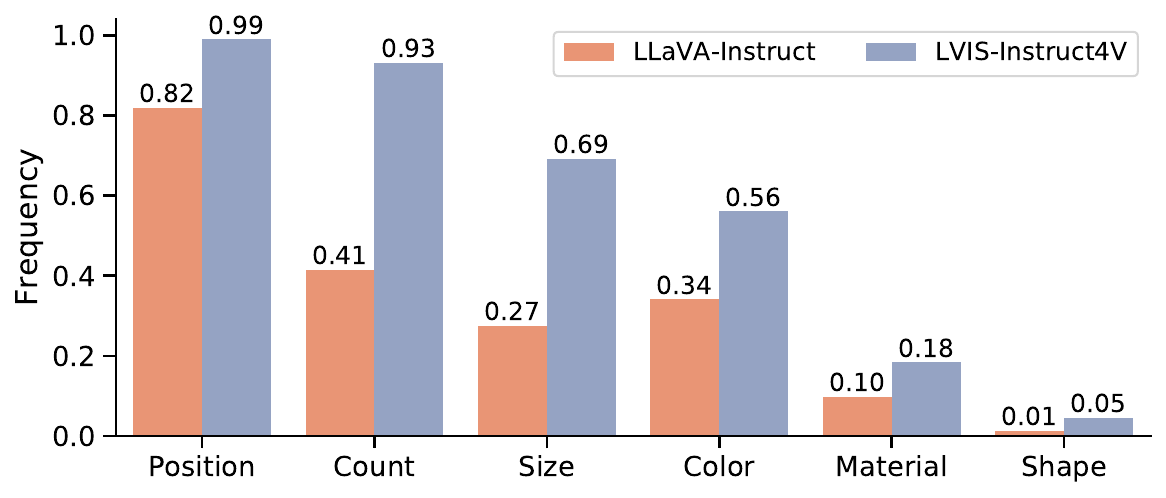}
  \vspace{-0.1in}
  \caption{Statistics of instruction data of LLaVA-Instruct and  LVIS-Instruct4V.}
  \label{fig:instruct_analysis}
\end{figure}

\fakeparagraph{Instruction analysis.} To gain a better understanding of instructions derived with and without visual inputs, we investigate the distribution of words related to ``position'', ``count'', ``size'' ``color'', ``material'', and ``shape'', which are fine-grained clues, for LLaVA-Instruct and LVIS-Instruct4V. We then instantiate each clue with a list of words. For example, words like ``black'', ``blue'', ``brown'', ``gray'', \etc are used as indicators of color. Please refer to Appendix~\ref{appendix:vocab_list} for the complete list. We then
calculate the occurrence of words within the instructions. The resulting counts are normalized by the total number of instructions. Figure~\ref{fig:instruct_analysis} shows the results. We can observe that instructions from \dataname indeed exhibit a greater richness in terms of fine-grained information.

\section{Experiment}
\label{sec:exp}
\subsection{Implementation Details}
We adopt the same model architectures with LLaVA-1.5~\cite{liu2023improved}. By default, we mix the conversational data in \dataname with the benchmark data, a collection of task-related data~\cite{goyal2017vqav2,okvqa,hudson2019gqa,mishra2019ocrvqa,schwenk2022okvqa,singh2019textvqa,kazemzadeh2014referitgame,krishna2017visual,sharegpt}, following~\cite{liu2023improved}.  We also additionally incorporate the descriptional data in \dataname and LLaVA-Instruct for data scaling experiments. The batch size, image resolution, and training schedules are also the same as LLaVA-1.5. 

\subsection{Comparison with State-of-the-art Methods}

\begin{table}[t!]
\caption{Comparison with state-of-the-art LMMs on various benchmarks. 
$^\ddag$ indicates incorporating the descriptional data in \dataname.
w/ G: use GPT models to generate instruction-following data, Priv: the data are private. $^{*}$ denotes the training images of the datasets are observed during training. V-7/13B indicates Vicuna-7/13B~\cite{vicuna}; L-7/13B indicates LLaMA-7/13B~\cite{touvron2023llama,touvron2023llama2}. Q-7B indicates Qwen-7B~\cite{qwen-report}. The best results are marked in bold.}
\renewcommand\arraystretch{1.2}
\centering
    \begin{subtable}{\linewidth}
    \centering
    \caption{Results on traditional VQA benchmarks.}
    \footnotesize
    \scalebox{0.90}{
    \begin{tabular}{p{24mm} p{10mm} p{16mm} | p{8mm}p{8mm}p{8mm}p{8mm}p{8mm}  }
    \toprule
    Method & LLM & w/ G & VQA$^\text{v2}$ & GQA & VisWiz & SQA$^\text{I}$ & VQA$^\text{T}$\\
    \midrule
    InstructBLIP~\cite{dai2023instructblip} & {\footnotesize V-7B} & GPT-4 & -- & 49.2 & 34.5 & 60.5 &  50.1  \\
    IDEFICS-9B~\cite{idefics} & {\footnotesize L-7B} & \XSolidBrush & 50.9 & 38.4 & 35.5 & – & 25.9 \\
    Qwen-VL~\cite{bai2023qwen} & {\footnotesize Q-7B} & Priv & 78.8$^*$ & 59.3$^*$ & 35.2 & 67.1 & \textbf{63.8} \\
    Qwen-VL-chat~\cite{bai2023qwen} & {\footnotesize Q-7B} & Priv & 78.2$^*$ & 57.5$^*$ & 38.9 & 68.2 & 61.5 \\
    LLaVA-1.5~\cite{liu2023improved} & {\footnotesize V-7B} & GPT-4 & 78.5$^*$ & 62.0$^*$ & 50.0 & 66.8 & 58.2 \\
    \rowcolor{Gray}
    \system & {\footnotesize V-7B} & GPT-4V & 79.2$^*$ & 62.6$^*$ & 52.5 & 68.4 & 57.6 \\
    \rowcolor{Gray}
    \system$^\ddag$-mixLLaVA & {\footnotesize V-7B} & GPT-4 \& 4V & 79.6$^*$ & 62.6$^*$ & 51.8 & 68.3 & 58.7 \\
    \midrule
    BLIP-2~\cite{li2023blip} & {\footnotesize V-13B} & \XSolidBrush & 41.0 & 41.0 & 19.6 & 61.0 & 42.5 \\
    InstructBLIP~\cite{dai2023instructblip} & {\footnotesize V-13B} & GPT-4 & -- & 49.5 & 33.4 & 63.1 & 50.7\\
    IDEFICS-80B~\cite{idefics} & {\footnotesize L-65B}& \XSolidBrush & 60.0 & 45.2 & 36.0 & -- & 30.9  \\
    Shikra~\cite{chen2023shikra} & {\footnotesize V-13B} & GPT-4 & 77.4$^*$ & -- & -- & -- & -- \\
    LLaVA-1.5~\cite{liu2023improved} & {\footnotesize V-13B} & GPT-4 & 80.0$^*$ & 63.3$^*$ & 53.6 & \textbf{71.6} & 61.3 \\
    \rowcolor{Gray}
    \system & {\footnotesize V-13B} & GPT-4V & 80.1$^*$ & \textbf{63.8}$^*$ & 51.4 & 69.0 & 62.1 \\
    \rowcolor{Gray}
    \system$^\ddag$-mixLLaVA & {\footnotesize V-13B} & GPT-4 \& 4V & \textbf{80.7}$^*$ & 63.6$^*$ & \textbf{57.2} & 70.6 & 62.5 \\
    \bottomrule
    \end{tabular}}
    \label{b:benchmark_vlbench}
    \end{subtable}
    \begin{subtable}{\linewidth}
    \centering
    \caption{Results on recent LMM benchmarks.}
    \footnotesize
    \scalebox{0.90}{
    \begin{tabular}{p{24mm} p{10mm} p{16mm}|p{8mm} p{8mm}p{8mm} p{10mm} p{10mm} l }
    \toprule
    Method & LLM & w/ G & POPE & MME & MMB  & SEED & LLaVA$^\text{W}$ & MM-Vet \\
    \midrule
    InstructBLIP~\cite{dai2023instructblip} & {\footnotesize V-7B} & GPT-4 & -- & -- & 36.0  & 53.4 & 60.9 & 26.2 \\
    IDEFICS-9B~\cite{idefics} & {\footnotesize L-7B} & \XSolidBrush & -- &  -- & 48.2  & -- & -- & -- \\
    Qwen-VL~\cite{bai2023qwen} & {\footnotesize Q-7B} & Priv & -- & -- & 38.2  & 56.3 & -- & -- \\
    Qwen-VL-chat~\cite{bai2023qwen} & {\footnotesize Q-7B} & Priv & -- & 1487.5 & 60.6  & 58.2 & -- & -- \\
    LLaVA-1.5~\cite{liu2023improved} & {\footnotesize V-7B} & GPT-4 & 85.9 & 1510.7 & 64.3  & 58.6 & 63.4 & 30.5 \\ 
    \rowcolor{Gray}
    \system & {\footnotesize V-7B} & GPT-4V & 84.0 & 1472.9	& 67.1  & 60.8 & 70.4 & 34.6 \\
    \rowcolor{Gray}
    \system-mixLLaVA & {\footnotesize V-7B} & GPT-4 \& 4V & \textbf{86.0} & 1528.2 & 66.2  & 60.6 &	67.0 & 31.5 \\
    \midrule
    BLIP-2~\cite{li2023blip} & {\footnotesize V-13B} & \XSolidBrush &  85.3 & 1293.8 & --  & 46.4 & 38.1 & 22.4 \\
    InstructBLIP~\cite{dai2023instructblip} & {\footnotesize V-13B} & GPT-4 & 78.9 & 1212.8 & --  & -- & 58.2 & 25.6 \\
    IDEFICS-80B~\cite{idefics} & {\footnotesize L-65B}& \XSolidBrush & -- & -- & 54.5  & -- & -- & -- \\
    Shikra~\cite{chen2023shikra} & {\footnotesize V-13B} & GPT-4 & -- & -- & 58.8  & -- & -- & -- \\
    LLaVA-1.5~\cite{liu2023improved} & {\footnotesize V-13B} & GPT-4  & 85.9 & 1531.3 & 67.7 & 61.6 & 70.7 & 35.4 \\
    \midrule
    \rowcolor{Gray}
    \system & {\footnotesize V-13B} & GPT-4V & 85.3 & 1572.0 & 67.8 & \textbf{62.5} & \textbf{76.7} & \textbf{40.2} \\
    \rowcolor{Gray}
    \system$^\ddag$-mixLLaVA & {\footnotesize V-13B} & GPT-4 \& 4V & \textbf{86.0} & \textbf{1574.9} & \textbf{68.0}  & 61.6 & 71.3 & 37.4 \\
    \bottomrule
    \end{tabular}}
    \label{tab:benchmark_llmbench}
    \end{subtable}
\label{tab:benchmark}
\end{table}

In Table~\ref{tab:benchmark}, we present a comprehensive comparison between the model trained using our data, and other large multimodal models (LMMs) across a diverse array of benchmarks. Using Vicuna-7B as the language model, our method achieves 79.2 on VQAv2~\cite{goyal2017vqav2}. When the size of the language model is scaled to 13B, the result is further improved to 80.1. Notably, our 7B and 13B models significantly outperform the counterparts of LLaVA-1.5 models across various recently proposed challenging LMM benchmarks, \eg, LLaVA$^w$ and MM-Vet, by only changing the LLaVA-Instruct to our \dataname.

In order to further boost the performance, we additionally use LLaVA-Instruct during training, and the resulting model achieves state-of-the-art results on the majority of these benchmarks, specifically, 9 out of 11. Remarkably, employing the same architecture and training procedure, it surpasses LLaVA~\cite{liu2023improved} across all benchmarks. We achieve a substantial 43.6 gain (1574.9 \vs 1531.3, measured by scores) on MME~\cite{fu2023mme} and an impressive 5.6\% improvement on MM-Vet~\cite{yu2023mmvet}. It is worth pointing out that MME and MM-Vet are challenging benchmarks designed to evaluate LMMs, which contain fine-grained multimodal tasks like counting, OCR, \etc. These results highlight the effectiveness of \dataname.

\subsection{Discussion}
\noindent \textbf{Ablation studies.} We study the effects of optimization and training schedules on instruction tuning in Table~\ref{tab:ablation}. The results in rows 1 and 2 show that even if the language model is kept frozen during instruction finetuning (Ours-13B-mixLLaVA-fzLLM), our data could still improve the performance on LMM benchmarks like MME and SEED compared with LLaVA-1.5, validating its superior quality. In addition, the performance on MME could be boosted to 1581.3 if we finetune the LLaVA-1.5 model on our data (Ours-13B-mixLLaVA-ftLLaVA-1.5), surpassing its original score by 50.0 (1531.3 \vs 1581.3). Finally, we also find that training the model for a longer schedule (Ours-13B-mixLLaVA-2epoch) could not bring additional performance gains, and training for one epoch as LLaVA-1.5 leads to overall better results.

\begin{table}[!ht]
\centering
\renewcommand\arraystretch{1.2}
\small
\caption{Ablation studies on both traditional VQA benchmarks and LMM benchmarks.}
\label{tab:ablation}
\scalebox{0.90}{
\begin{tabular}{lp{8mm}p{8mm}p{8mm}p{8mm}l}
\toprule
Method & VQA$^\text{v2}$ & GQA & VisWiz & MME & SEED \\
\midrule
LLaVA-1.5-13B & 80.0 & 63.3 & 53.6 & 1531.3 & 61.6 \\
%Ours-13B & 80.1 & 51.4 & 73.0 & 1572.0 & 62.5 \\ 
\midrule
Ours-13B-mixLLaVA-fzLLM & 80.0 & 62.9 & 50.0 & 1544.3 & \textbf{61.7} \\
Ours-13B-mixLLaVA-ftLLaVA-1.5 & 80.6 & 63.8 & 51.9 & \textbf{1581.3} & 61.6 \\
Ours-13B-mixLLaVA-2epoch & 80.5 & \textbf{64.0} & 56.3 & 1538.4 & 61.6 \\
Ours-13B-mixLLaVA & \textbf{80.7} & 63.6 & \textbf{57.2} & 1574.9 & 61.6 \\
\bottomrule
\end{tabular}}
\end{table}

\vspace{0.05in}
\noindent \textbf{Visualizations.} We now provide qualitative visualization of results produced by different models in Table~\ref{tab:tricky_example}. We can see that by additionally training with \dataname, our model is better at recognizing small texts and identifying the relationships among objects.

\begin{table}[!h]
\caption{Our model achieves more accurate scene text recognition and finer-grained differentiation.}
\begin{minipage}{0.4\textwidth}
\centering
\scalebox{0.88}{
\begin{tabular}{l p{3.5cm} }
\toprule
 \multicolumn{2}{l}{\bf Scene text recognition:}  \\
\midrule
\multicolumn{2}{r}{\includegraphics[width=4cm]{imgs/scene_text_2.jpg}} \\
\midrule
User & What are the scene texts in the image? \\
\ \\
 &  \\
\ \\
\ \\
\ \\
\ \\
\midrule
LLaVA-1.5 & The scene text in the image reads ``Capilano Suspension Bridge.'' \\
\midrule
\system & The scene texts in the image are ``Capilano Suspension Bridge'' and ``\textbf{Be astonished by the natural beauty}.'' \\
\bottomrule
\end{tabular}
}
\end{minipage}
\ 
\begin{minipage}{0.6\textwidth}
\scalebox{0.88}{
\begin{tabular}{l p{6.5cm} }
\toprule
 \multicolumn{2}{l}{\bf Fine-grained visual differentiation:}  \\
\midrule
\multicolumn{2}{r}{\includegraphics[width=6cm]{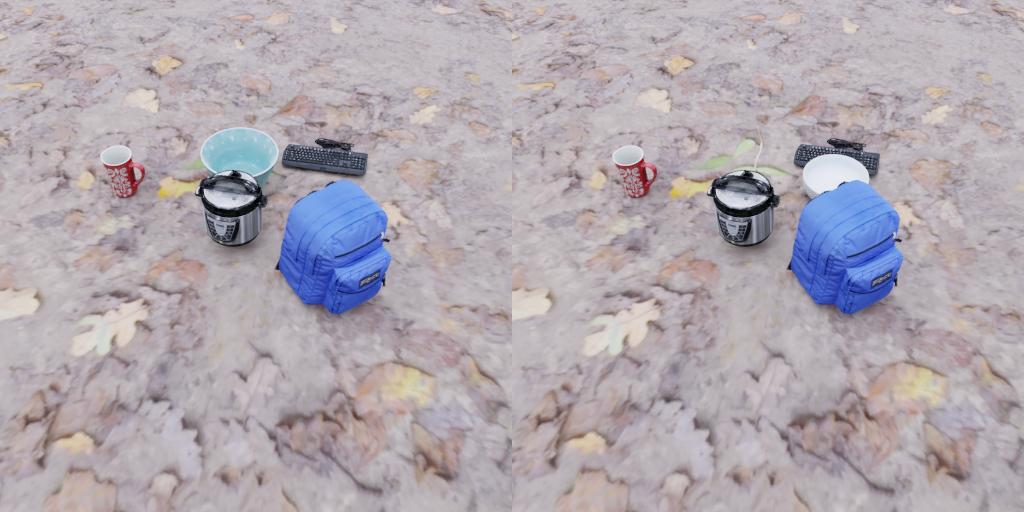} } \\
\midrule
User & There are two images (left and right). Now you have two captions: CaptionA: ``the \textbf{blue bowl} is placed on the right of the small red coffee mug'', and CaptionB: ``the \textbf{white bowl} is on the right of the small red coffee mug''. Please indicate which caption corresponds to the left image and which caption corresponds to the right image.\\
\midrule
LLaVA-1.5 & CaptionA corresponds to the right image and CaptionB corresponds to the left image.\\
\ \\
\midrule
\system & CaptionA corresponds to the left image and CaptionB corresponds to the right image.\\
\ \\
\ \\
\ \\
\bottomrule
\end{tabular}
}
  \end{minipage}
    \label{tab:tricky_example}
\vspace{-0.1in}
\end{table}

\section{Conclusion}
\label{sec:conclusion}
Existing instructions for training large multimodal models are typically generated by prompting language models with textual descriptions, which are derived from image annotations. While offering decent results, without ever seeing images, the performance of these instructions is upper-bounded by manual annotations in current benchmarks, which exist in the forms of captions and bounding boxes.  The reasons are two-fold: (1) these annotations are often times coarse-grained; (2) LLMs can generate inaccurate answers without the entire visual context. To address these challenges, we introduced a large-scale visual instruction dataset \dataname, which contains $220K$ fine-grained and visually aligned instructions. \dataname is curated by harnessing the great visual analysis ability of GPT-4V through carefully designed prompts. We then built upon LLaVA, a state-of-the-art of multimodal language model, to evaluate the efficacy of \dataname. We conducted extensive experiments on a wide range of benchmarks and the results clearly demonstrate that the proposed dataset benefits visual instruction tuning.

\bibliographystyle{abbrv}
\bibliography{main}

\appendix
\onecolumn
\section{Pseudo code for instruction data generation}
\label{appendix:prompt}
\begin{algorithm*}[!ht]
  \caption{Pseudo code for instruction data generation with GPT-4V~\cite{gpt4v}.}
  \label{alg:gpt4v}
\var{PROMPT\_DICT}\{ \\
    \KwSty{prompt\_conversation}: 
    ( \; \hspace{2mm}``You are an AI visual assistant, and you are seeing a single image. Answer all questions as you are seeing the image. Design a conversation between you and a person asking about this photo. The answers should be in a tone that a visual AI assistant is seeing the image and answering the question. Ask diverse questions and give corresponding answers. Include questions asking about the visual content of the image, including the object types, counting the objects, object actions, object locations, relative positions between objects, etc. Only include questions that have definite answers: (1) one can see the content in the image that the question asks about and can answer confidently; (2) one can determine confidently from the image that it is not in the image. Do not ask any questions that cannot be answered confidently. Also include complex questions that are relevant to the content in the image, for example, asking about background knowledge of the objects in the image, asking to discuss events happening in the image, etc. Again, do not ask about uncertain details, but the questions should be challenging enough, requiring the person to utilize 1) complex reasoning; 2) world knowledge; 3) explanatory answers; and 4) multi-turn conversation, to give accurate answers. Please provide detailed answers when answering complex questions. For example, give detailed examples or reasoning steps to make the content more convincing and well-organized. Please ensure all the questions are closely related to the visual content of the provided image, which means that if the person cannot see the picture but only gets access to the text description of it, he/she will not be able to answer accurately. If the AI assistant asks counterfactual questions, the person should give a negative answer, rather than making up an answer.'' \;
    ), \; 
    \KwSty{prompt\_detail}: 
    ( \; \hspace{2mm}``You are an AI visual assistant that can analyze a single image. You will receive several sets of descriptions of the objects in the image, each in the format: set id, category of the 1st object: location of the 1st object; category of the 2nd object: location of the 2nd object; .... Note that different sets are separated by a link break. These locations are in the form of bounding boxes, represented as (x1, y1, x2, y2) with floating numbers ranging from 0 to 1. These values correspond to the top left x, top left y, bottom right x, and bottom right y. Utilize the supplied bounding box details to formulate a precise and comprehensive caption that accurately describes the highlighted objects within their designated areas, regardless of the objects\' sequence. Note that please describe the overall information of all objects in each set, rather than giving each object in a set an independent description. To generate high-quality captions, you should first understand the visual content, then based on the background knowledge or reasoning, either explain why the things are happening that way or provide guidance to facilitate the better understanding of these objects. Ensure that the caption is tightly connected to the objects in question, meticulously avoiding any reference to extraneous objects beyond the defined locations.  If addressing the query accurately becomes too challenging, please respond with [failed]. Instead of directly mentioning the bounding box coordinates, utilize this data to explain the scene using natural language.  Include details like object counts, the position of the objects, relative position between the objects.   When using the information from the categories and coordinates, directly explain the scene, and do not mention that the information source is the caption or the bounding box. Always answer as if you are directly looking at the image. Please proceed with the tasks in the set order, respectively. Make sure that the caption for each set begins with "Bbox List", followed by a set id. The object information about categories and normalized coordinates are as follows: \ArgSty{Category, (x1, y1, x2, y2); Category, (x1, y1, x2, y2); ... }'' \;) \; \}
 
\KwSty{output} = \FuncCall{openai.ChatCompletion.create}{ \;
\hspace{13mm}     \var{model="gpt-4v"}, \;
\hspace{5mm}  \var{messages=[{"role": "user", "content": \KwSty{Image}; \KwSty{prompt}} ]}, \;
    }
\end{algorithm*}

\section{Vocabulary List} \label{appendix:vocab_list}

We define a list of vocabularies for instruction analysis.

\begin{table*}[h]
\small
\centering
\setlength{\tabcolsep}{8.pt}
\renewcommand{\arraystretch}{1.2}
% @{\extracolsep{\fill}}
\begin{tabular*}{\linewidth}{l|p{11cm}}
\toprule
\textbf{Type} & \textbf{Words} \\
\midrule
Position & `above', `ahead', `ahead position', `back', `back direction', `backward', `behind', `behind position', `below', `down', `downward', `downward direction', `elevated', `elevation', `forward', `forward direction', `front', `front position', `frontward', `higher', `left', `left direction', `leftward', `lower', `lower position', `more ahead', `more back', `more behind', `more forward', `more front', `more left', `more right', `on', `on position', `overhead', `overhead position', `right', `right direction', `rightward', `underneath', `underneath position', `up', `upward', `upward direction'\\
\midrule
Count & `one', `two', `three', `four', `five', `six', `seven', `eight', `nine', `ten'\\
\midrule
Size & `average', `averageness', `big', `bigness', `considerably', `enormity', `enormous', `giant', `giantness', `huge', `hugeness', `large', `largeness', `larger', `little', `littleness', `massive', `massiveness', `medium', `medium size', `middling', `mini', `miniature', `minuteness', `moderate', `moderately', `moderateness', `more average', `more enormous', `more middling', `more miniature', `more moderate', `petite', `petiteness', `slightly', `small', `smaller', `smallness', `tininess', `tiny'\\
\midrule
Color & `black', `blue', `brown', `gray', `green', `orange', `pink', `purple', `red', `white', `yellow'\\
\midrule
Material & `cotton', `glass', `glassy', `iron', `leather', `linen', `linen-like', `metal', `plastic', `polymeric', `satiny', `silk', `steel', `stone', `stony', `timber', `velvet', `wooden', `woody'\\
\midrule
Shape & `circle', `circular', `hexagon', `hexagonal', `octagon', `octagonal', `oval', `rectangle', `rectangular', `square', `triangle', `triangular'\\
\bottomrule
\end{tabular*}
\caption{List of words for instruction analysis. }
\label{tab:wordtype}
\end{table*}

\end{document}